\documentclass[conference]{IEEEtran}
\IEEEoverridecommandlockouts
\usepackage{cite}
\usepackage{amsmath,amssymb,amsfonts}
\usepackage{algorithmic}
\usepackage{graphicx}
\usepackage{textcomp}
\usepackage{xcolor}

\usepackage{url}
\usepackage{subfigure}

\def\BibTeX{{\rm B\kern-.05em{\sc i\kern-.025em b}\kern-.08em
    T\kern-.1667em\lower.7ex\hbox{E}\kern-.125emX}}
\begin{document}

\title{Wind Estimation in Unmanned Aerial Vehicles with Causal Machine Learning}

\author{
\IEEEauthorblockN{Abdulaziz Alwalan}
\IEEEauthorblockA{\textit{School of Aerospace, Transport and Manufacturing} \\
\textit{Cranfield University}\\
United Kingdom\\
aziz.walan@gmail.com}

\and
\IEEEauthorblockN{Miguel Arana-Catania*}
\IEEEauthorblockA{\textit{School of Aerospace, Transport and Manufacturing} \\
\textit{Cranfield University}\\
United Kingdom \\
miguel.aranacatania@cranfield.ac.uk}
*Corresponding author
}

\maketitle

\begin{abstract}
In this work we demonstrate the possibility of estimating the wind environment of a UAV without specialised sensors, using only the UAV's trajectory, applying a causal machine learning approach. We implement the causal curiosity method which combines machine learning times series classification and clustering with a causal framework. We analyse three distinct wind environments: constant wind, shear wind, and turbulence, and explore different optimisation strategies for optimal UAV manoeuvres to estimate the wind conditions. The proposed approach can be used to design optimal trajectories in challenging weather conditions, and to avoid specialised sensors that add to the UAV's weight and compromise its functionality.
\end{abstract}


\section{Introduction}
Multi-rotor Unmanned Aerial Vehicles (UAVs) have become increasingly popular in commercial and research sectors due to advantages over fixed-wing UAVs, such as vertical take-off and landing, hovering capabilities, and the ability to yaw on the spot. However, a significant challenge for those UAVs is their vulnerability to wind disturbances, which can affect their flight stability and energy efficiency \cite{abichandani2020wind}. Thus, recognising the effects of wind and incorporating this understanding into flight controls can enhance safety and efficiency.

Currently, several methods allow different types of UAVs to measure wind speed. For instance, fixed-wing UAVs can gauge wind speed using sensors like the pitot tube, while multi-rotors can employ sonic anemometers. Nonetheless, these methods come with various constraints. For instance, it is essential to place flow sensors at a distance from the rotor's turbulence. This arrangement might be straightforward for fixed-wing UAVs, but it poses challenges for multi-rotors. Another limitation of utilising a dedicated wind sensor is that it consumes a portion of the UAV's mass budget, potentially compromising the inclusion of other components.

To overcome these limitations, in this article we propose the use of a machine learning approach capable of identifying environmental conditions such as wind using only the UAV's position information. This approach, called causal curiosity \cite{sontakke2020causal}, is set in the framework of causal machine learning. Unlike traditional machine learning approaches that focus mainly on identifying patterns and correlations in the data, one of the objectives of this framework is to distinguish between correlation and causation and establish and use the causal relationships between variables. In this particular case, identifying the relationship between the wind condition, the cause, and changes in the UAV's trajectory, the effect. In this way, using UAV trajectory data, we can identify the specific wind conditions, without the need for specific sensors to measure the wind.

The original proposal of this method \cite{sontakke2020causal} was demonstrated in the case of a robotic agent interacting with different objects in its environment. The study enabled a robotic fingers agent to conduct experiments that assist in classifying the interaction with unknown objects and consequently infer the properties of those objects. These parameters that determine the causal dynamics of the interactions are called causal factors. They are parameters such that, by applying a certain sequence of actions on the environment, the observations obtained are organised in distinguishable disjoint sets according to the parameter values. For example, in the latter case, the mass and friction of the objects determined how they moved when lifted or pushed by the robotic hand. Causal curiosity enables agents to discover optimal action sequences to identify causal factors that influence the environment's dynamics. Through this, the agents learn to deduce a categorisation or representation for the actual causal factors in every environment they are placed in. The strategies found by the agents possess logical meaning, such as learning to lift blocks in order to classify them by weight. Therefore, we could use the same causal curiosity reward system on a position trajectory of a UAV moving in different wind environments allowing us to identify wind conditions without the need for a specific sensor for wind speed measurement.

In this article we make the following contributions:
\begin{itemize}
\item We apply for the first time the causal curiosity approach to the aerospace domain and show that it is effective in identifying wind conditions for UAVs using only their position data. 
\item We analyse the effectiveness of the method in different types of wind conditions (constant wind, shear wind, and turbulence) and in terms of its main parameters. 
\item We evaluate the results of using an optimisation methodology for the optimal action selection.
\end{itemize}

\section{Related work}

In recent years, UAVs have become an indispensable tool in many industries such as logistics and communications and their use is predicted to keep growing further in the future \cite{MOURTZIS2021183}. Nevertheless, an essential aspect of drone delivery optimisation is energy consumption. Energy usage during flight, especially hovering while loaded, is substantial. An essential factor affecting energy consumption during the flight is weather, especially wind conditions. Tailwinds can accelerate a drone, while headwinds can decelerate it \cite{ito2022load}.
Moreover, energy limitation due to size and weight constraints is a significant challenge for UAVs in many applications such as communication systems \cite{zeng2017energy}. Understanding external factors such as wind conditions is crucial, impacting its ability to maintain specific trajectories and affecting its overall energy efficiency.

\subsection{UAVs wind conditions identification methods}
Over time, the field of wind and airspeed measurement in fixed-wing small UAVs has undergone extensive exploration and enhancement. These efforts have predominantly revolved around employing flow sensors, specifically pitot tubes, combined with the dynamic models intrinsic to these UAVs. An exemplary study \cite{cho2011wind}, utilised an extended Kalman filter technique combined with the Wind Triangle Relation to deduce horizontal wind velocities and their directions. This methodology also integrated a calibration factor for pitot tube readings to gauge airspeed accurately \cite{abichandani2020wind}.

Further research \cite{johansen2015estimation} aimed to determine wind speed, Angle-Of-Attack (AOA), and Sideslip Angle (SSA) for a fixed-wing UAV using kinematic relationships combined with a Kalman Filter from the UAV's airspeed and attitude data. This approach eliminates the need for detailed aerodynamic models or specific aircraft data. The system works efficiently if the UAV frequently changes its attitude during flight. While the results for wind speed align well with ground observations, the AOA and SSA estimates need further validation but show promising correlations.

Another approach \cite{borup2016nonlinear}, devised a nonlinear wind observer. By merging standard aircraft models with readings from devices such as the GPS, IMU, and pitot tube, this method could estimate wind speed and drone airspeed.

Rounding off these innovations, \cite{borup2019machine} introduces an approach to calculate air parameters data for a small UAV MEMS-based pressure sensor placed on the UAV's exterior. These data inform a machine-learning model that predicts the UAV's AOA, SSA, and airspeed. Two machine learning techniques - artificial neural networks and linear regression - were evaluated using data from wind tunnel tests and real flights. Training with wind tunnel data presented accuracy challenges, while real flight data proved more reliable.

Fixed-wing UAVs move in the direction they face because of their design and have a non-zero turning radius. This makes it straightforward to measure airspeed with external sensors, as their structure offers convenient spots for mounting these sensors, typically on the nose or wings, without significant disruption from rotor turbulence. In contrast, mounting airspeed or wind speed sensors on multi-rotor UAVs is more difficult. Their design does not provide ample space for these sensors. Moreover, measurements from these sensors can be distorted by rotor turbulence and other effects caused by the rotors. Additionally, these sensor systems must consider that multi-rotor UAVs have a non-zero turning radius but can swiftly move in any given direction \cite{abichandani2020wind}.

However, there are other ways of airspeed measurement. For example, in \cite{thielicke2021towards} a lightweight drone was crafted to carry a high-precision sonic anemometer. To ensure high accuracy in dynamic conditions, it is crucial to use a full-size anemometer, maintain a considerable distance between the anemometer and the propellers, and deploy a robust algorithm to minimise propeller-induced airflow effects. In practical tests, the drone successfully gauged wind velocity in a wind turbine's wake, with its results closely matching lidar data and theoretical predictions.

Other techniques take advantage of existing onboard sensors. The algorithm in \cite{perozzi2022using} is designed to utilise an inertial measurement unit (IMU) combined with an earth reference tracking system and sensors that monitor rotor speed. Three distinct algorithms were tested through numerical experiments on a nonlinear quadrotor simulator. The algorithms effectively estimated airspeed, particularly on the horizontal plane.

Upon examining the diverse methods available for UAV wind condition identification, it becomes evident that each method, while innovative in its approach, has inherent limitations. Traditional sensor-based methods, while often precise, impose design and weight constraints on the UAV. Fixed-wing UAVs provide more convenient options for mounting such sensors away from disruptive factors like rotor turbulence. However, multi-rotors present a challenge due to their design intricacies, which do not afford many optimal placements for these sensors without interference.

Furthermore, using dedicated wind sensors can compromise the UAV's weight allocation, potentially necessitating trade-offs with other vital components. In \cite{perozzi2022using} the authors developed wind estimation algorithms which do not require additional sensors, but they are tailored to quadrotors, with a foundational basis on their detailed aerodynamic model. While the research provides valuable insights and advancements for quadrotor applications, its specificity can also be seen as a limitation. This focused approach means that the methodologies and conclusions derived might not be directly transferable or applicable to other types of drones or aerial vehicles. Diverse aerial platforms, such as hexacopters, octocopters, or fixed-wing drones, have distinct aerodynamic properties and dynamics. As a result, any wind estimation method that specifically caters to quadrotors may not address the unique challenges and nuances posed by these other aerial vehicles. Thus, the need for broader, more encompassing solutions for varied drone architectures remains evident. As the applications and complexities of UAVs continue to grow, it is imperative to devise methods that can circumvent these challenges and be universally applicable.

In light of these challenges, our decision to employ a different approach becomes justified. Drawing inspiration from  \cite{sontakke2020causal}, we advocate for a shift towards a machine learning based method which does not rely on a specific wind sensor, and instead capitalises on the existing sensors onboard the UAV such as GPS. By adopting this method, not only can we alleviate the challenges posed by traditional methods, but also harness the vast potential that such an approach can offer in enhancing UAV performance in varied wind conditions.

\subsection{Causal curiosity in machine learning}
In this work we apply a methodology set within the framework of causal machine learning. This combines ideas from causal analysis, focused on understanding the causal relationship between the different variables of a system, with machine learning, in this case focused on learning which actions are the most optimal for the previous causal analysis.

In relation to causal analysis, this is a field solidly established in several works and theoretical formulations such as \cite{pearl2000models,glymour2016causal,peters2017elements,spirtes2000causation}. It is an analysis rooted in the field of statistics and nevertheless despite this basis has only recently started to connect with the field of machine learning. The combination of the two in what is becoming known as Causal Machine Learning presents different presentations and foci of interest. \cite{scholkopf2022causality} offers a historical introduction to the field culminating with a presentation of different areas of machine learning in which causal analysis can be relevant and a special focus on causal representation learning. This last point is developed in more detail by \cite{scholkopf2021toward}. Causal representation learning focuses on the representation of the systems in which learning takes place and its connection with the causal relationships between the elements of the system. Representation is a key issue for machine learning whether we are talking about the initial external representation of the data used or the latent internal representation produced by training our models. These representations are often disconnected from the causal approach and therefore from the variables in which it would be natural to represent the system in order to understand its dynamics. Identifying and working with these variables can greatly increase the efficiency, generability and interpretability of our machine learning models. \cite{kaddour2022causal} provides a very broad and detailed overview of the intersection between machine learning and causal analysis and of the challenges found.

Other works \cite{bareinboimcrlonline,zeng2023survey,grimbly2021causal} focus in particular on causal reinforcement learning. In general, reinforcement learning works with variables that do not necessarily have a causal relationship between them, and therefore are less effective in solving certain tasks, such as estimating the value of some of them from the effect produced by the variables in our control.

The methodology we apply in this work \cite{sontakke2020causal} is in turn connected to the concept of curiosity. Curiosity in robotics is seen as an inner reward that pushes a reinforcement learning agent to explore unknown parts of its environment in connection with the exploration-exploitation dilemma. Several authors have explored the concept of curiosity in robotics \cite{pathak2017curiosity,burda2018large,schmidhuber1991curious,chentanez2004intrinsically,
oudeyer2009intrinsic,still2012information, baldassarre2013intrinsically, baranes2013active, barto2013intrinsic,mohamed2015variational,forestier2017intrinsically, laversanne2018curiosity, oudeyer2018computational}. However, in this case the concept of curiosity reward is linked to the identification of different causal factors, through the exploration of actions that produce an adequate differentiation between environments with different values, connecting thus it in a direct way with the concepts coming from causal analysis. The most recent work on causal curiosity in robotics applies the methodology to the case of planetary exploration \cite{mcdonnell2024autonomous}.

\section{System model}

In this section, we describe the system we have simulated in this project. We designed a computer simulation of a UAV with a specific weight and cross-sectional area, navigating through distinct wind scenarios for a given time.

The drone interactions with the wind are modelled and parametrised by its mass $m$, drag coefficient $C_D$, cross-sectional area $S$ and air density $\rho$. The forces on the drone are calculated at every time step of the simulation using the following force equations:

\begin{align}
    \vec{F}_{net} &= \vec{F}_{thrust} + \vec{F}_{drag} + \vec{F}_{gravity} \label{eq 2} \\
    \vec{F}_{drag} &= -\frac{1}{2} \rho \vec{v}_{air}^2 C_D   S \label{eq 3} \\
    \vec{F}_{gravity} &= -9.81m \label{eq 4}
\end{align}

The dynamics of the drone are calculated using the following equations of motion:
\begin{align}
    \vec{a} &= \frac{\vec{F}_{net}}{m} \label{eq 5} \\
    \vec{v}_{air} &= \vec{v}_{ground} + \vec{v}_{wind} \label{eq 6} \\
    \vec{v}_{air} &= \vec{v}_{o,air} + \vec{a}\, dt \label{eq 7} \\
    \vec{r} &= \vec{r}_{o} + \vec{v}_{ground}\, dt \label{eq 8}
\end{align}

The wind conditions considered are constant wind, shear wind, and turbulent wind. For constant wind, the value of $\vec{v}_{wind}$ stays constant for the duration of the simulation. For shear wind, the value of $\vec{v}_{wind}$ is calculated using the power law wind-shear model \cite{elkinton2006investigation} using the following equation:
\begin{equation}
    \frac{U(z)}{U(z_{r})} = \left(\frac{z}{z_{r}}\right)^\alpha \label{eq 9}
\end{equation}

where $U(z)$ is the wind speed at altitude $z$, and $z_{r}$ is the relative altitude.
Finally, for the turbulent wind, the Dryden wind turbulence
model is applied, using the implementation of \cite{abichandani2020wind}. This implementation uses the Dryden transfer function defined in \cite{moorhouse1982background} for the X, Y, and Z directions as follows:
\begin{align}
    H_{x}(s) &= \sigma_{x} \sqrt{\frac{2L_{x}}{\pi v_{x}}} \, \frac{1}{1+\frac{L_{x}}{v_{x}}s} \label{eq 10} \\
    H_{y}(s) &= \sigma_{y} \sqrt{\frac{L_{y}}{\pi v_{y}}} \, \frac{1+ \frac{\sqrt{3}L_{y}}{v_{y}}s}{\left(1+\frac{L_{y}}{v_{y}}s\right)^2} \label{eq 11} \\
    H_{z}(s) &= \sigma_{z} \sqrt{\frac{L_{z}}{\pi v_{z}}} \, \frac{1+ \frac{\sqrt{3}L_{z}}{v_{z}}s}{\left(1+\frac{L_{z}}{v_{z}}s\right)^2} \label{eq 12}
\end{align}

\begin{figure}[!ht]
    \centering
    \subfigure[UAV trajectory (left) and wind direction and magnitude on multiple trajectory points (right).]{
    \includegraphics[width=0.48\textwidth]{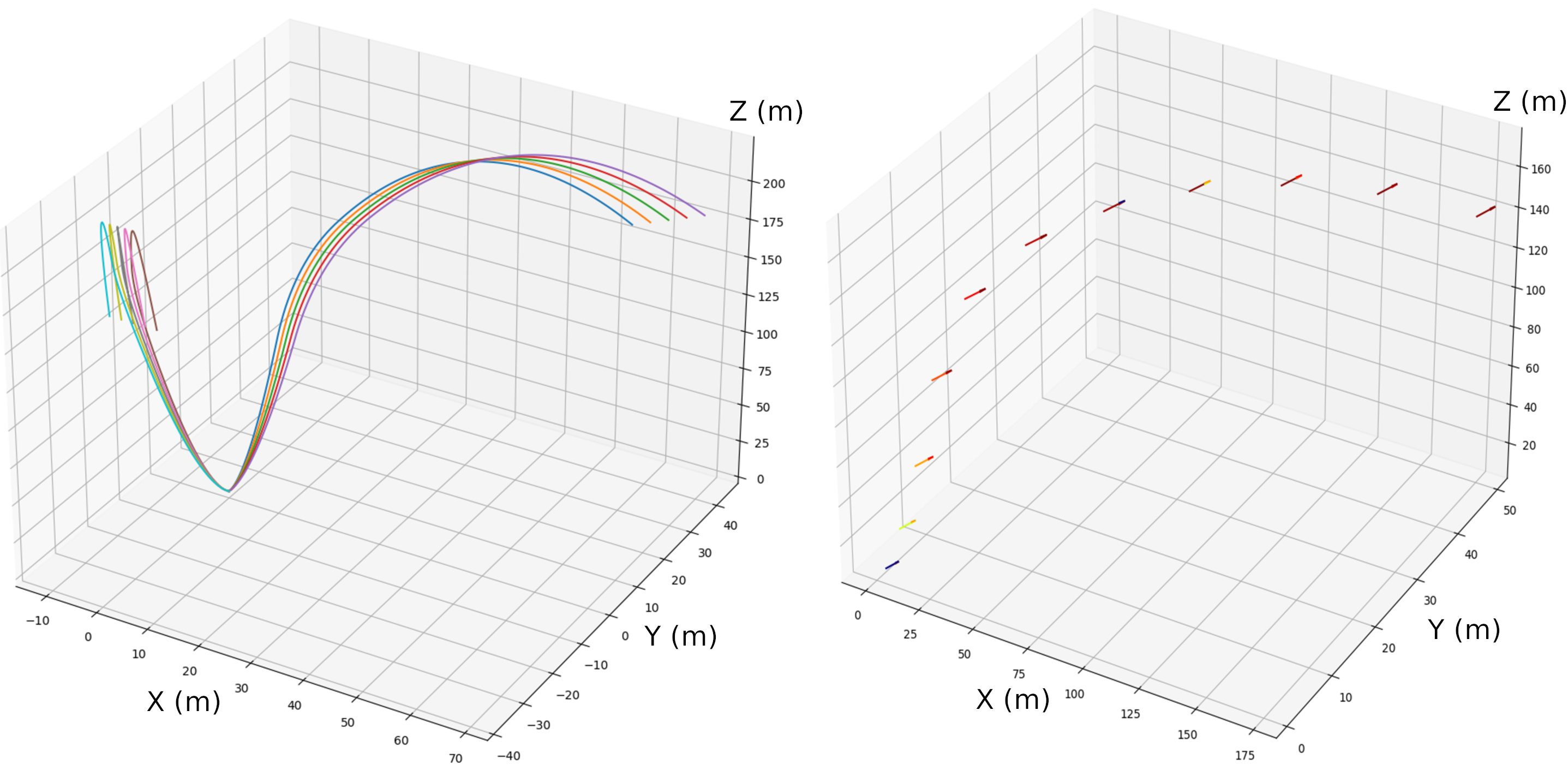}}
    \subfigure[UAV velocity direction and magnitude along the trajectory (left) and wind direction and magnitude on multiple trajectory points (right).]{
    \includegraphics[width=0.48\textwidth]{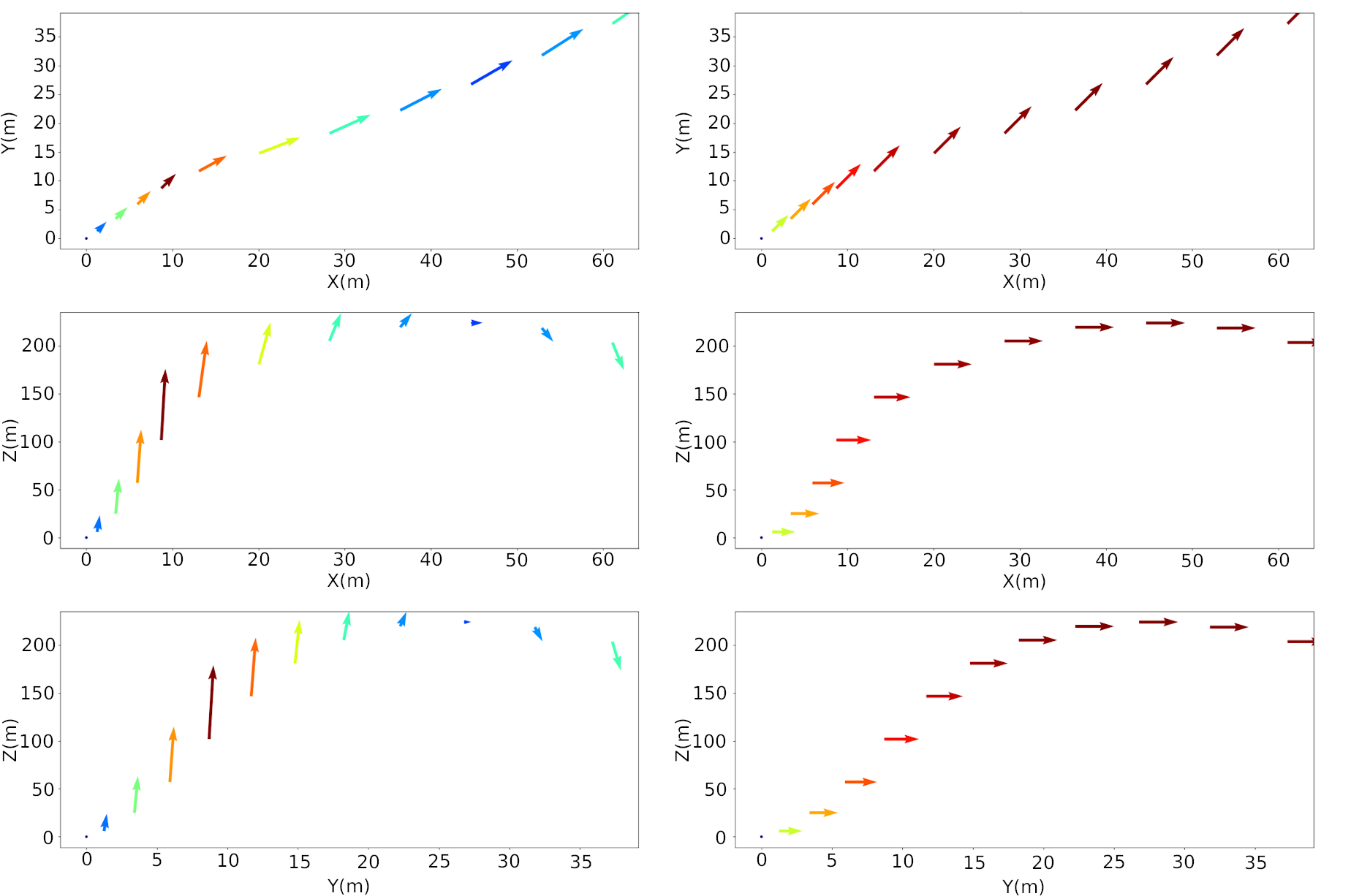}}
    \caption{UAV position and wind velocities through the trajectory.}
    \label{fig:pos_trajectory}
\end{figure}

The simulation is represented in Figure \ref{fig:pos_trajectory}. The left plot in (a) shows the UAV position trajectory during the simulation of two wind environments.
The right plot in (a) shows the wind speed and direction at the trajectory of the UAV in one of the wind environments. For this latter case, the plots in (b) show the UAV and wind velocities represented in 2D perspectives XY, XZ, and YZ.

\section{Methodology}
\label{sec:method}

The aim of our methodology is to distinguish between different wind conditions based on the trajectories. To do so, we use machine learning time series classification to identify the trajectories and assign them to each wind condition.

In order to compare the trajectories we use Dynamic Time Warping (DTW) \cite{muller2007dynamic}. This is a method that allows for flexible comparisons between two time series, potentially of different lengths. The core idea behind DTW is to find the optimal alignment between two time series that minimises the distance (or ``warping cost") between them. This is achieved by ``warping" the time dimension of one or both series to get an optimal match. The strength of DTW lies in its ability to identify patterns in sequences that might be out-of-sync or have different temporal stretches. For example, if a UAV follows a particular trajectory pattern but at varying speeds in different instances, DTW can capture the inherent pattern across these varied speeds, making it invaluable for tasks like UAV trajectory classifications.

For the clustering process we use k-means clustering. In the context of time series classification, k-means works by partitioning time series data into k number of clusters based on their similarity. Each cluster's centroid represents a mean value for the time series data points within that cluster.

The implementation of these methods has been done using the tslearn\footnote{~\url{https://tslearn.readthedocs.io}} library \cite{JMLR:v21:20-091}, an extension of the widely used scikit-learn library \cite{pedregosa2011scikit} specialised for time series data tasks. 

To gauge the quality of clusters created by the clustering algorithm we use the Silhouette score metric. The score ranges from -1 to 1, where a high value indicates that the object is well-matched to its own group and poorly matched to neighbouring groups. In the context of our UAV trajectory classification, the Silhouette score becomes essential to understand how distinct our time series clusters are. A good Silhouette score would indicate that trajectories within the same cluster are similar, while trajectories across different clusters are distinct. \cite{shahapure2020cluster} offers an in-depth exploration of the Silhouette score and its use in cluster quality evaluation.

According to \cite{sontakke2020causal}, some actions done by the agent could provide more useful information about causal factors than other actions. For example, in the robotic scenario analysed in that work, if the robotic fingers agent is tasked with classifying an object by its weight, a lifting action is more useful than a pushing action on the object. The reason is that friction and weight can cause resistance when pushing. In our case, some UAV actions (i.e. thrust sequences) could give more information about wind conditions, and other UAV actions can give less information. For example, if the UAV is tasked with classifying if it is flying in shear wind or constant wind, actions (or thrust) in the vertical plane are more useful than the ones in the horizontal plane because shear wind speed varies with altitude while constant wind speed does not vary. In our experiment we test an optimisation strategy to find these optimal sequences of actions. 

\begin{figure}[!ht]
    \centering
    \subfigure[System overview using random exploration.]{
    \includegraphics[width=0.47\textwidth]{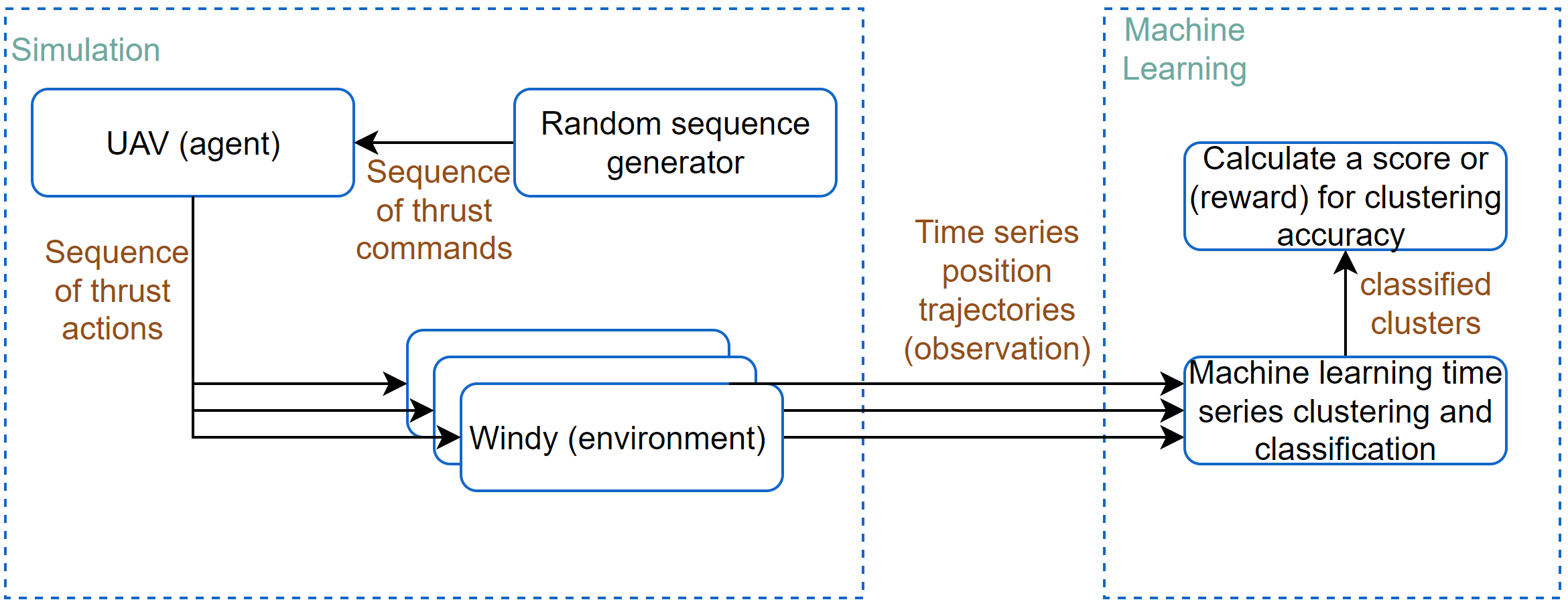}}
    \subfigure[System overview using CEM.]{
    \includegraphics[width=0.47\textwidth]{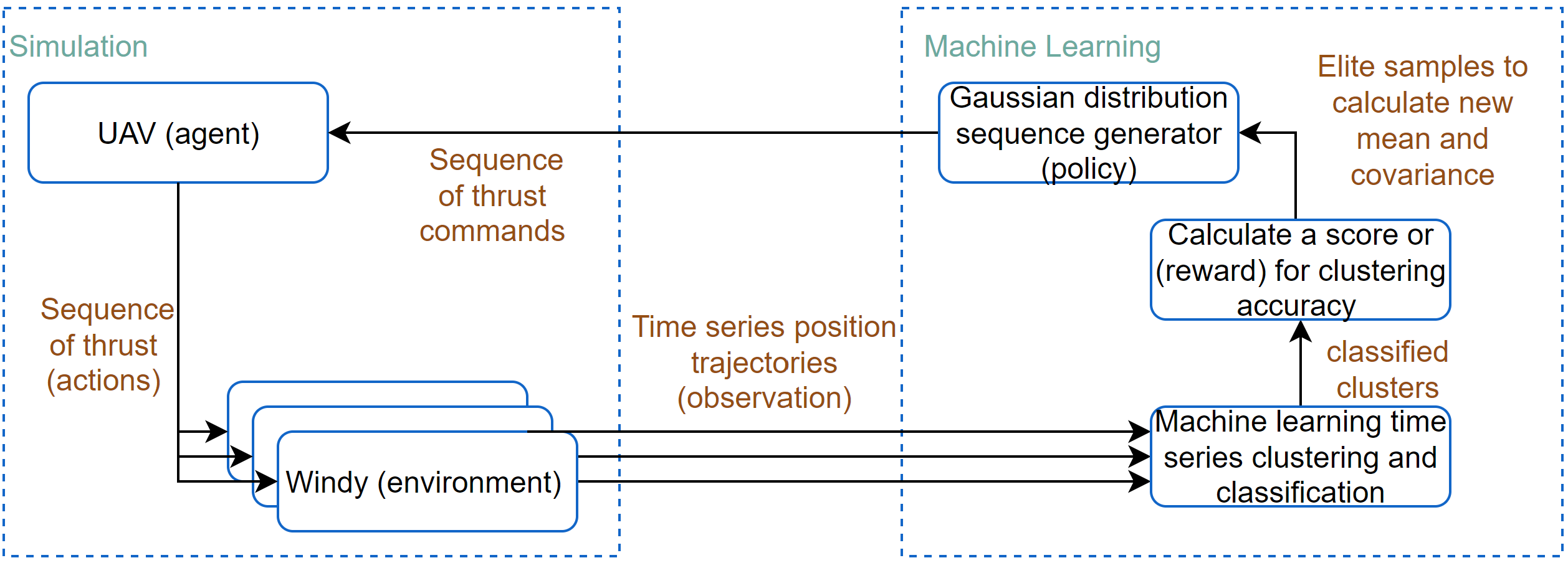}}
    \caption{System flow charts using different optimal action search strategies.}
    \label{fig:sys_flow}
\end{figure}

The system's overall framework is depicted in Figure \ref{fig:sys_flow}(a), where the simulation begins with a UAV executing consistent thrust actions in varied wind conditions in two different wind environments. Once completed, the UAV's position trajectories of the two environments, being time series, are compared using the Dynamic Time Warping algorithm and clustered using the k-means technique. The final classification receives a Silhouette score. The objective of this machine learning model is to differentiate between the wind conditions by examining the UAV's position trajectories. Since certain actions yield more precise classifications, it is crucial to identify action strategies that optimise the classification. Additionally, the cross-entropy method (CEM) was utilised as a search strategy, as shown in Figure \ref{fig:sys_flow}(b).
 
The CEM method uses the following algorithm \cite{botev2013cross}:
\begin{enumerate}
    \item Initialise: Choose $\hat{\mu}_0$ and $\hat{\sigma}_0^{2}$. Set t = 1.
    \item Draw: Generate a random sample $X_1, . . ., X_N$ from the $N(\hat{\mu}_{t-1},\hat{\sigma}_{t-1}^{2})$ distribution where $X_i=(X_{i,1},...,X_{i,n})$
    \item Select: Let $I$ be the indices of the $N_e$ best performing (=elite) samples. 
    \item Update: For all $j = 1, ..., n$ let
    
\end{enumerate}
\begin{equation}
    \hat{\mu}_{t,j} = \sum_{k\in I} X_{k,j}/N_e \label{eq CEM1}
\end{equation}
\begin{equation}
    \hat{\sigma}_{t,j}^{2} = \sum_{k\in I} (X_{k,j}-\hat{\mu}_{t,j})^2/N_e \label{eq CEM2}
\end{equation}

In this method, a multivariate Gaussian distribution probability with mean $\mu$ and standard deviation $\sigma$ is used to produce $N$ samples. In this case, the samples represent thrust magnitude values. Next, the simulation is run for 1 loop for each one of the $N$ samples. Based on the Silhouette score, the $N_e$ best-performing samples are used to calculate a new mean and standard deviation. These cycle back into the same algorithm until no improvement is shown. See an example of this approach in \cite{kroese2006cross}.

\section{Experiments}
\label{sec: Experimentation}

In our experiments, we evaluated how different thrust sequences give different information about the wind conditions. Initially, we used a random exploration strategy to analyse the effect of the parameters that define the environment and the interaction with it in the information obtained. The parameters explored are wind speed, range length, the total number of wind condition environments, the number of thrust changes, and wind direction. 

In each experiment the objective is to differentiate between two types of wind conditions. For example, the first can be light shear wind speeds, and the second light constant wind speeds. For each of the two conditions, we create 5 environments that represent variations over that wind condition. The UAV interacts with the 10 environments carrying out the same sequence of thrust but describing different trajectories depending on the surrounding wind.  

The simulation starts with the UAV having a pre-specified random thrust magnitude and direction, and the thrust vector value changes six times at pre-set times $t_{1}, t_{2},...,t_{6}$. After running the simulation, the ten UAV trajectories are clustered and classified using Soft Dynamic Time Warping and the k-means method. Finally, these clusters are given a Silhouette score. This process constitutes a single simulation loop. The random sequence generation tests ran for 50 loops except if specified otherwise.

In Experiment 1 we explored the effect of the wind conditions in the classification. The pairs of wind conditions that were used in the experiments are described in Table \ref{tab:wind speed_conditions}.

\begin{table}[h!]
    \centering
    \caption{Wind conditions experimented.}
    \begin{tabular}{| c | c | c |} 
    \hline
    Scenario & Wind condition 1  & Wind condition 2 \\ 
    \hline\hline
        1 & light constant   & light constant  \\
        2 & light constant   & strong constant  \\
        3 & strong constant   & strong constant  \\
        4 & light shear   & light shear  \\
        5 & light shear   & strong shear  \\
        6 & strong shear   & strong shear  \\
        7 & light constant   & strong turbulence \\
        \hline
    \end{tabular}
    \label{tab:wind speed_conditions}
\end{table}

Experiment 1 allowed us to understand the effect of different combinations of wind types. After analysing the results, it was found that several of the scenarios were qualitatively similar, and thus the rest of the experiments were performed with a representative selection of the previous combinations.

The objective of Experiment 2 was to test the effect of the gap size between the two groups' wind speeds on the classification. The goal of Experiment 3 was to test the effect of the number of environments on the classification. The test was done using 10, 8, 6, 4, and 3 environments for each wind condition. Experiment 4 was performed to test the effect of the number of thrust vector changes on the classification. The test was done with 12, 10, 8, 6, 4, 2, and 0  thrust vector changes. The wind conditions used in Experiments 2, 3 and 4 are the scenarios 1 to 4 shown in Table \ref{tab:wind speed_conditions}. 

The objective of Experiment 5 was to test the effect of wind direction on classification. Therefore, all parameters like wind speed, range length, the total number of wind condition environments, and the number of thrust vector modifications remained unchanged. In one of the two groups, the wind speed sign was altered to negative. The testing was done on light constant wind vs light shear wind conditions.

These previous experiments analyse the possibility of identifying different combinations of wind conditions and the effect of the parameters of the environments and the interactions within them. Experiment 6 uses a search strategy to optimise the search for optimal thrust sequences. The strategy used in this case is the cross-entropy method, described in Section \ref{sec:method}. 

 To exemplify the experiments, we show here the details in the case of the scenario in which we try to distinguish between strong shear wind and light constant wind. Light wind speeds are in the range of 0.45-1.34 m/s, and strong wind speeds are in the range of 11.18-13.86 m/s according to the US National Weather Service\footnote{~\url{https://www.weather.gov/pqr/wind}}. For each of the wind conditions, we generate 5 environments, with wind parameters shown in Table \ref{tab:sim_env_input}. We note that the shear wind speeds are $U(z_r)$ in equation \ref{eq 9} at ground altitude or $z_r =0 $.

\begin{table}[h!]
    \centering
    \caption{Parameters of the wind models for the classification of strong shear wind vs light constant wind environments.}
    \subfigure[Shear wind. $\alpha =0.143$, $\rho$ = 1.225 Wind speed for different axes.]{
    \begin{tabular}{|p{1cm}|p{1cm}|p{1cm}|}    
        \hline
        x axis (m/s) & y axis (m/s) & z axis (m/s) \\
            \hline    \hline
        2.1 & 10.1 & 0 \\
        2.2 & 10.2 & 0 \\
        2.3 & 10.3 & 0 \\
        2.4 & 10.4 & 0 \\
        2.5 & 10.5 & 0 \\ 
            \hline
    \end{tabular}}
    \subfigure[Constant wind. $\rho$ = 1.225 ~~~~~~~~Wind speed for different axes.]{
    \begin{tabular}{|p{1cm}|p{1cm}|p{1cm}|}
        \hline
        x axis (m/s) & y axis (m/s) & z axis (m/s) \\
            \hline    \hline
        1.1 & 1.1 & 0 \\
        1.2 & 1.2 & 0 \\
        1.3 & 1.3 & 0 \\
        1.4 & 1.4 & 0 \\
        1.5 & 1.5 & 0 \\ 
            \hline
    \end{tabular}}
    \label{tab:sim_env_input}
\end{table}
  
  The UAV used in all the experiments is defined with the mass, drag coefficient, and cross-sectional area shown in Table \ref{tab:sim_sys_input}. These properties represent an average micro UAV available in global markets such as the 1.38 Kg DJI Phantom 4 or the 3.44 Kg DJI Inspire 2 \cite{singhal2018unmanned}. Additional simulation parameters are also shown: time step, total time, and number of loops. In these experiments, the thrust vector and time marks for when the thrust force changes are generated randomly. However, thrust magnitude is in the range of 0 to 50 Newtons which is a reasonable upper limit since the max thrust for the DJI Phantom 4 is 36.6 N and 64 N for the DJI Inspire 2 \cite{kurbanov2020development,polukhin2022development}. 
  The current experiments are estimated to work with submeter accuracy in the position.

\begin{table}[h!]
    \centering
    \caption{Simulation and UAV parameters.}
    \subfigure[UAV properties]{
    \begin{tabular}{|c|c|}
    \hline
        UAV mass (Kg) & 2 \\
        $C_D$ & 0.1 \\
        $S$ ($m^2$) & 0.01 \\
    \hline
    \end{tabular}}
    \subfigure[simulation  properties]{
    \begin{tabular}{|c|c|}
    \hline
        Time step (s) & 0.1 \\
        Total time  (s) & 12 \\
        Number of loops  & 50 \\
    \hline
    \end{tabular}}
    \subfigure[UAV thrust action properties ]{
    \begin{tabular}{|c|c|}
    \hline
        $F_{thrust}$ magnitude (N) & Randomly selected\\
         & in the range [0-50]\\
        $F_{thrust}$ directions [x y z] & Randomly selected\\
        $F_{thrust}$ application times (s) & Randomly selected \\
    \hline
    \end{tabular}}
    \label{tab:sim_sys_input}
\end{table}

\begin{figure}[!ht]
    \centering
    \subfigure[UAV trajectories in two groups of 5 different wind environments.]{
    \includegraphics[width=0.23\textwidth]{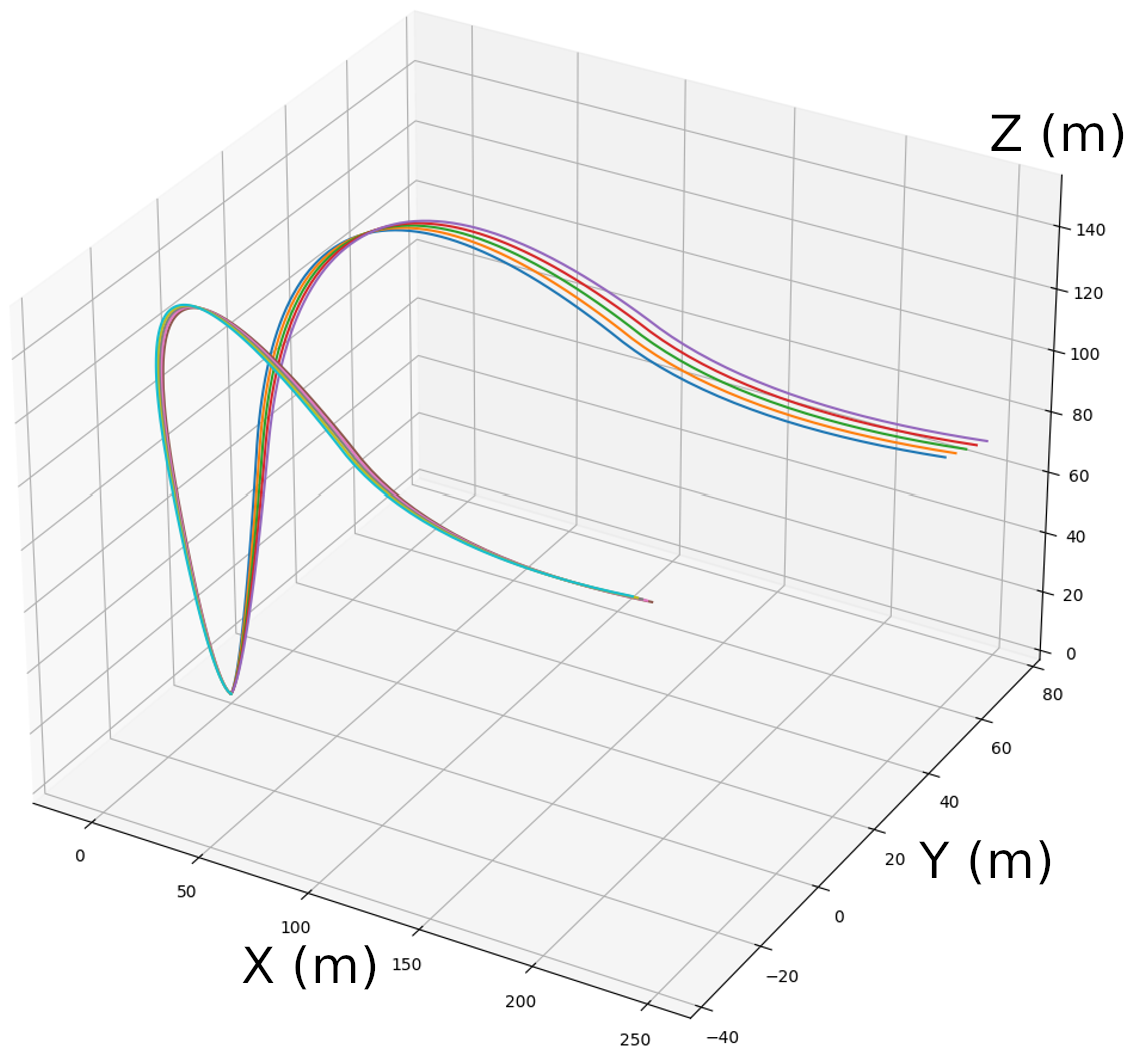}}
    \subfigure[Silhouette score for 50 loops.]{
    \includegraphics[width=0.23\textwidth]{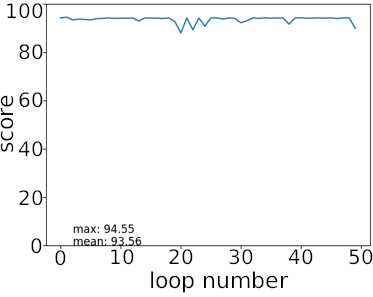}}
    \caption{Example trajectories for a single loop and scores for all of them.}
    \label{fig:sim_output}
\end{figure}

After one loop of simulations for two groups of 5 wind conditions is finished, UAV position trajectories are clustered and classified using the time series k-means and the Silhouette score is calculated. In Table \ref{tab:sim_output} can be seen the output of the simulation after finishing the simulation loops, showing the score, the classification of each trajectory in the 2 wind environment groups, the randomly generated time marks for thrust changes, and the randomly generated thrust vectors. The score of the loop of the example Figure \ref{fig:sim_output} (a) was 94.55. During the simulation, the UAV had a thrust of 40 Newtons in the z direction for 4 seconds, and then the thrust changed to 15N in the x direction for 6 seconds, and finally 45N in the y direction for 2 seconds. The maximum Silhouette score is 100. In this example, the observations are correctly classified into their corresponding groups. To visualise the trajectories, a plot is generated shown in Figure \ref{fig:sim_output} (a). The process is performed 50 times and plotted in (b), including the maximum score and mean score values.
\begin{table}[h!]
    \centering
    \caption{Simulation parameters and outputs after one simulation loop.}
    \begin{tabular}{|c|c|}
        \hline
        Silhouette score & 94.55 \\
        Observations1 & [1 1 1 1 1] \\
        Observations2 & [0 0 0 0 0] \\
        Thrust change times (s) & [ 0 4 10] \\
        Thrust magnitude (N) & [40 15 45] \\
        Thrust directions [x y z] & [[0 0 1] [1 0 0] [0 1 0]] \\
            \hline
    \end{tabular}
    \label{tab:sim_output}
\end{table}

It should be noted that when any trajectory is classified incorrectly in a certain loop, the Silhouette score in that loop is set to zero. 

\section{Results}
\subsubsection{Effect of Wind speed on classification}
\begin{figure}[!ht]
    \centering
    \subfigure[Light vs light constant winds.]{
    \includegraphics[width=0.23\textwidth]{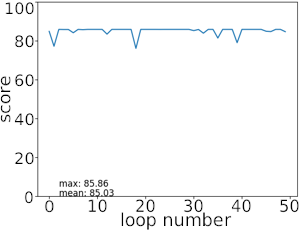}}
    \subfigure[Light vs strong constant winds.]{
    \includegraphics[width=0.23\textwidth]{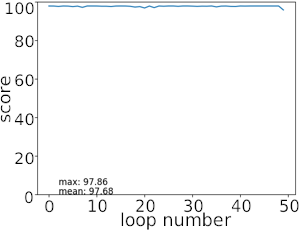}}
    \subfigure[Strong vs strong constant winds.]{
    \includegraphics[width=0.23\textwidth]{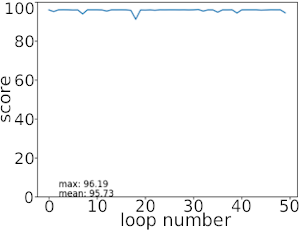}}
    \subfigure[Light vs light shear winds.]{
    \includegraphics[width=0.23\textwidth]{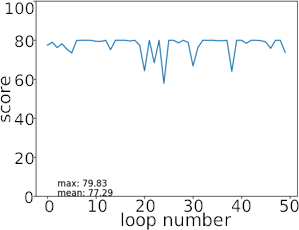}}
    \subfigure[Light vs strong shear winds.]{
    \includegraphics[width=0.23\textwidth]{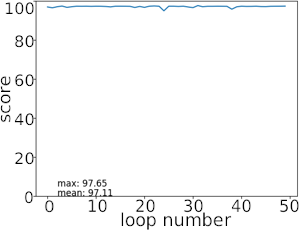}}
    \subfigure[Strong vs strong shear winds.]{
    \includegraphics[width=0.23\textwidth]{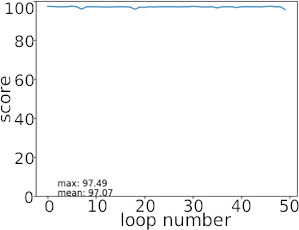}}
    \subfigure[Light constant wind vs strong turbulence.]{
    \includegraphics[width=0.23\textwidth]{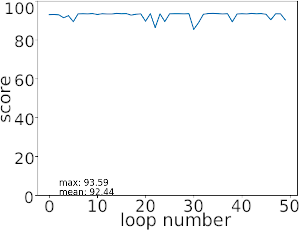}}
    \subfigure[Light constant wind vs light constant wind with 500 loops.]{
    \includegraphics[width=0.23\textwidth]{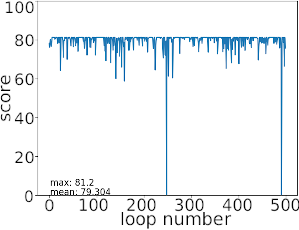}}
    \caption{Silhouette score vs loop number for the wind speed experiments.}
    \label{fig:wind speed}
\end{figure}

In this Experiment 1 we study the effect of the wind speed on the efficiency of the method. With the same steps done in the previous example, more experiments were done to compare wind speeds in both constant and shear wind models: light vs light, light vs strong, and strong vs strong. The results are shown in Figure \ref{fig:wind speed} (a) to (f). In all cases the average score was over 95 except for the two cases with only light wind conditions. In the first case, with light constant wind vs light constant wind, the maximum score was 85.86, and in the light shear wind vs light shear wind case, the maximum score was 79.83 as shown in Table \ref{tab:wind speed_output}. Additionally, these two cases had more frequent low scores in some of the loops (see the dips in the curves) which caused the mean to drop.

\begin{table}[!ht]
    \centering
    \caption{Maximum and mean Silhouette scores for all wind conditions scenarios in the wind speed experiments.}
    \begin{tabular}{|c|c|c|}
    \hline
    
    Scenario & Maximum & Mean\\
    \hline
    \hline 
        Light constant vs light constant & 85.86 & 85.03 \\
        Light constant vs strong constant & 97.86 & 97.68 \\
        Strong constant vs strong constant & 96.19 & 95.73 \\
        Light shear vs light shear & 79.83 & 77.29 \\
        Light shear vs strong shear & 97.65 & 97.11\\
        Strong shear vs strong shear & 97.49 & 97.07 \\
        Light constant vs strong turbulence & 93.59 & 92.44 \\
        Light constant vs light constant (500 loops) & 81.20 & 79.30 \\
\hline
    \end{tabular}
    \label{tab:wind speed_output}
\end{table}

An experiment with strong wind turbulence vs light constant wind was conducted. It is illustrated in Figure \ref{fig:wind speed} (g), the mean score is 92.44 and the maximum score is 93.59 which is slightly lower than most of the cases above. Nevertheless, it is higher than the two cases where both wind speeds were light. 

The last case analysed here was an experiment with 500 loops for the light constant wind vs light constant wind case. The result is shown in Figure \ref{fig:wind speed} (h). It can be seen how the maximum score did not improve despite extending the number of loops.

\subsubsection{Effect of range similarity on classification}
\begin{figure}[!ht]
    \centering
    \subfigure[Light vs light constant winds.]{
    \includegraphics[width=0.23\textwidth]{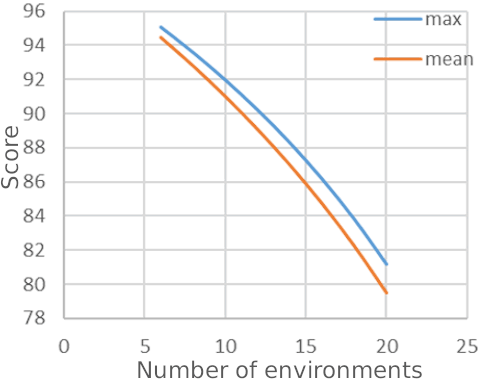}}
    \subfigure[Strong vs strong constant winds.]{
    \includegraphics[width=0.23\textwidth]{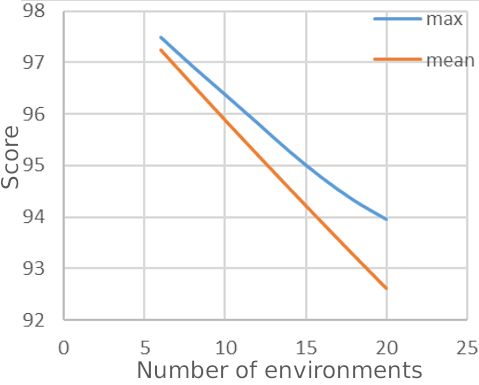}}
    \subfigure[Light vs strong constant winds.]{
    \includegraphics[width=0.23\textwidth]{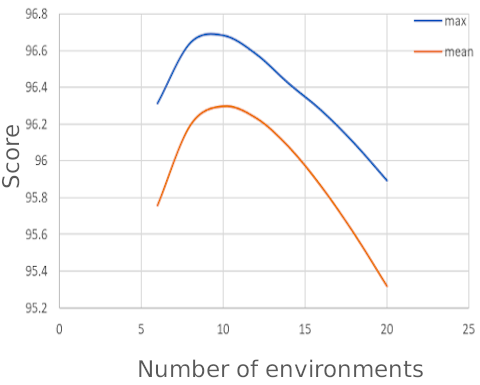}}
    \subfigure[Light vs light shear winds.]{
    \includegraphics[width=0.23\textwidth]{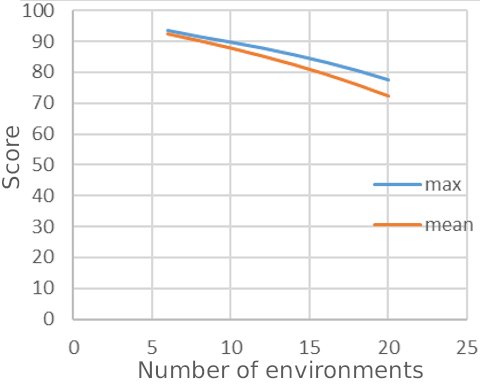}}
    \caption{Silhouette score vs the number of environments for the range similarity experiments. A decrease in the number of the environment is equivalent to an increase in the difference between the wind speed ranges of the two wind conditions.}
    \label{fig:Range length}
\end{figure}

In this section, we examine the impact of wind condition groups' similarities on the classification results. To discount the effect of changes in the average strength of the wind in this experiment, the length change in the study is done by eliminating the closest environments between the two groups. Starting with 10 wind environments for each group, the highest-value environment of the bottom group and the lowest-value environment of the top group are both removed, thus increasing the difference between the ranges. This is repeated until each group has 3 wind conditions environments out of the initial 10. Table \ref{tab:Range length_output} and Figure \ref{fig:Range length} show the result of experimentation on different cases, which generally show an increase in the score when the dissimilarity between the two groups is increased (meaning a decrease in the number of environments in the plots).

\begin{table}[!ht]
    \centering
    \caption{Maximum and mean Silhouette scores for all wind conditions scenarios in the range similarity experiments.}
    \begin{tabular}{|c|c|c|c|c|}
    \hline
    & \multicolumn{2}{ c| }{Maximum} &\multicolumn{2}{ c |}{Mean}\\
    \hline
    Scenario  & 6 env  &  20 env &  6 env &  20 env\\
    \hline 
        Light const. vs light const. & 95.05 & 81.19 & 94.45 & 79.50\\
        Light const. vs strong const. & 96.30 & 95.90 & 95.75  & 95.30 \\
        Strong const. vs strong const. & 97.47 & 93.96 & 97.25  & 92.61 \\
        Light shear vs light shear & 93.49 & 77.42 & 92.49 & 72.44  \\
        \hline
    \end{tabular}
    \label{tab:Range length_output}
\end{table}

\subsubsection{Effect of number of environments on classification}
\begin{figure}[!ht]
    \centering
    \subfigure[Light vs light constant winds.]{
    \includegraphics[width=0.23\textwidth]{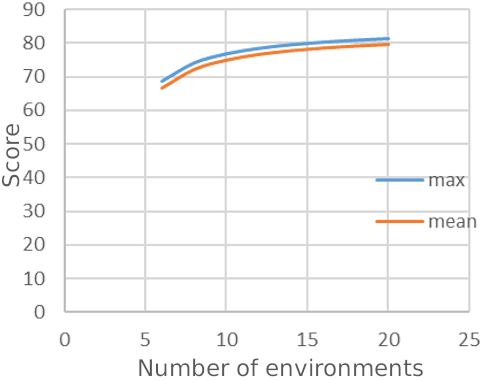}}
    \subfigure[Light vs strong constant winds.]{
    \includegraphics[width=0.23\textwidth]{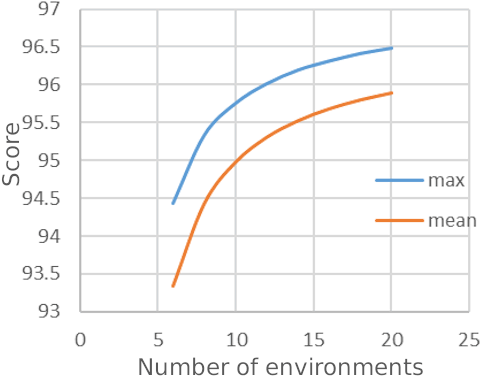}}
    \subfigure[Strong vs strong constant winds.]{
    \includegraphics[width=0.23\textwidth]{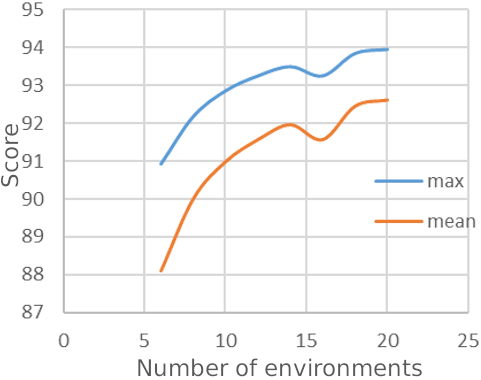}}
    \subfigure[Light vs light shear winds.]{
    \includegraphics[width=0.23\textwidth]{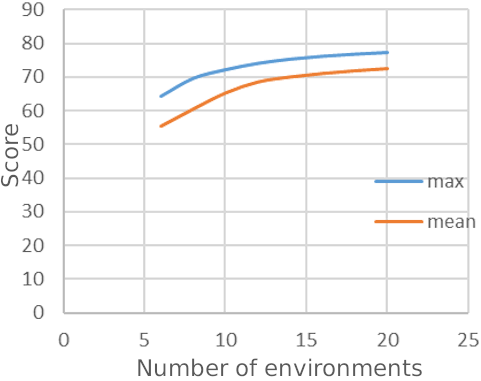}}
    \caption{Silhouette score vs number of environments. The number of environments changes but the difference between the wind speed ranges of the two wind conditions is constant.}
    \label{fig:Number of environments}
\end{figure}

In contrast with the previous experiment, here the distance between the wind speed ranges of the two wind conditions is kept constant while decreasing the number of environments. This is done by removing the environment with median wind speed in each group starting with 10 environments each and ending up with 3 environments each. Table \ref{tab:Number of environments} and Figure \ref{fig:Number of environments} show the results of the experiments. They show an increase in the score when the number of environments is increased. Another finding of this experiment is that the number of environments score improvement seems to plateau around 15 environments in each group.

\begin{table}[h!]
    \centering
    \caption{Maximum and mean Silhouette scores for all wind conditions scenarios in the number of environments experiments.}
    \begin{tabular}{|c|c|c|c|c|}
    \hline
    & \multicolumn{2}{ c| }{Maximum} &\multicolumn{2}{ c |}{Mean}\\
    \hline
    Scenario  & 6 env  &  20 env &  6 env &  20 env\\
    \hline 
        Light const. vs light const. & 68.80 & 81.19 & 66.46 & 79.50\\
        Light const. vs strong const. & 94.42 & 96.48 & 93.33  & 95.89 \\
        Strong const. vs strong const. & 90.91 & 93.96 & 88.10  & 92.61 \\
        Light shear vs light shear & 64.41 & 77.42 & 55.58 & 72.44  \\
        \hline
    \end{tabular}
    \label{tab:Number of environments}
\end{table}

\subsubsection{Effect of number of thrust changes on classification}
\begin{figure}[!ht]
    \centering
    \subfigure[Light vs light constant winds.]{
    \includegraphics[width=0.23\textwidth]{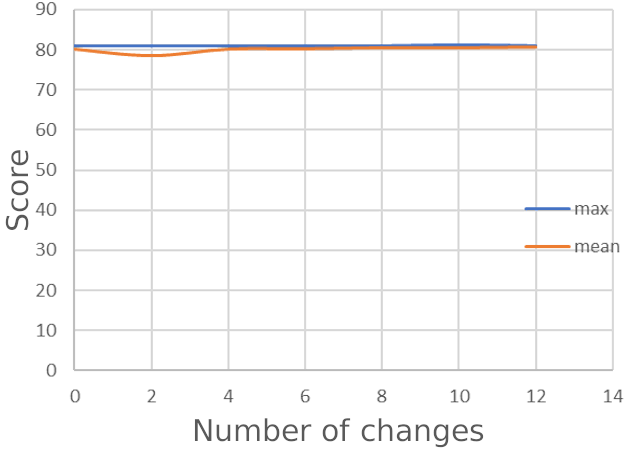}}
    \subfigure[Light vs strong constant winds.]{
    \includegraphics[width=0.23\textwidth]{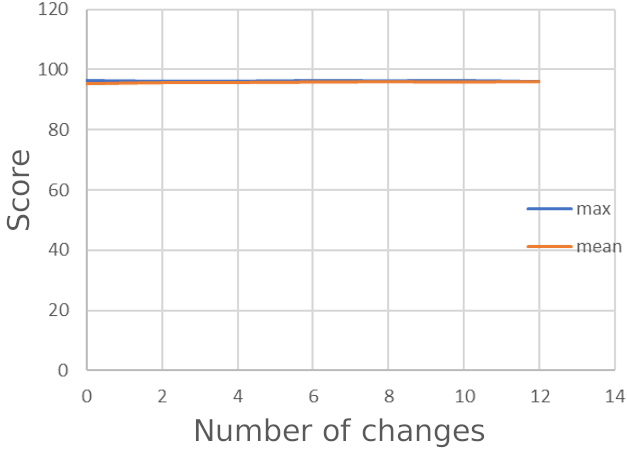}}
    \subfigure[Strong vs strong constant winds.]{
    \includegraphics[width=0.23\textwidth]{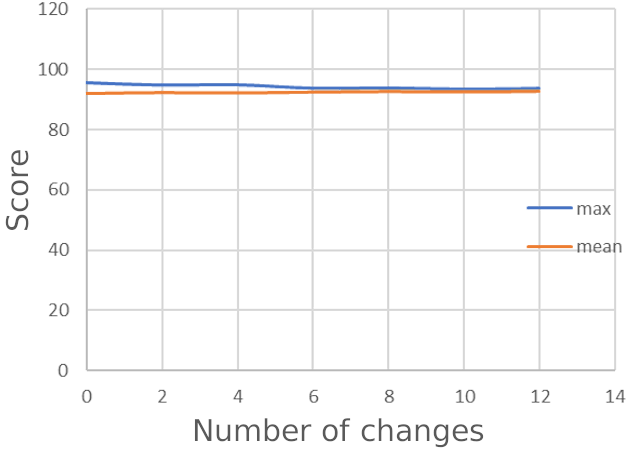}}
    \subfigure[Light vs light shear winds.]{
    \includegraphics[width=0.23\textwidth]{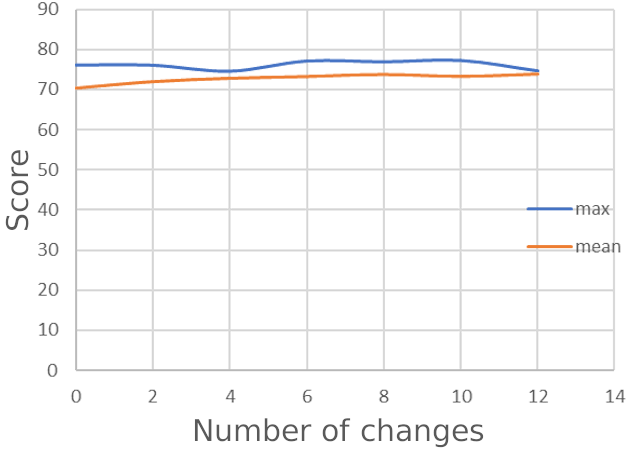}}
    \caption{Silhouette score vs number of UAV's thrust changes during the simulation.}
    \label{fig:thrust changes}
\end{figure}

In this test, both range distance and number of environments are kept constant. Here, the number of UAV thrust vector changes during the simulation is varied starting with no change and ending with 12 changes which means a change every second. The results of the experiment are presented in Table \ref{tab:thrust changes} and Figure \ref{fig:thrust changes}. They do not show any major change in the score.

\begin{table}[h!]
    \centering
    \caption{Maximum and mean Silhouette scores for all wind conditions scenarios in thrust changes experiments. 0 and 12 changes.}
    \begin{tabular}{|c|c|c|c|c|}
    \hline
    & \multicolumn{2}{ c| }{Maximum} &\multicolumn{2}{ |c |}{Mean}\\
    \hline
    Scenario  & 0 ch.  &  12 ch. &  0 ch. &  12 ch.\\
    \hline 
        Light const. vs light const. & 81.19 & 81.19 & 80.11 & 80.61\\
        Light const. vs strong const. & 96.55 & 96.21 & 95.51  & 96.09 \\
        Strong const. vs strong const. & 95.52 & 93.67 & 92.03  & 92.81 \\
        Light shear vs light shear & 76.17 & 74.70 & 70.26 & 73.88  \\
        \hline
    \end{tabular}
    \label{tab:thrust changes}
\end{table}

\subsubsection{Effect of wind direction on classification}
\begin{figure}[!ht]
    \centering
    \subfigure[Light constant vs light shear wind.  Similar direction.]{
    \includegraphics[width=0.23\textwidth]{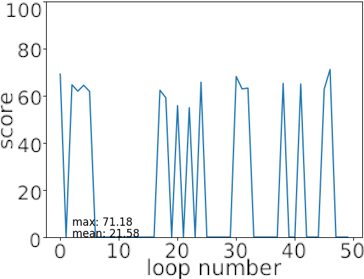}}
    \subfigure[Light constant vs light shear wind. Opposite direction.]{
    \includegraphics[width=0.23\textwidth]{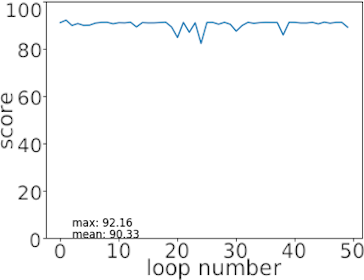}}
    \caption{Silhouette score vs loop number where the two groups of wind conditions had (a) similar or (b) opposite wind direction.}
    \label{fig:wind direction}
\end{figure}

In this case, all other parameters (wind speed, range length, number of wind condition environments, and number of thrust vector changes) are kept constant while the wind speed sign was changed to negative in one of the two groups. Table \ref{tab:wind direction_output} and Figure \ref{fig:wind direction} show two cases of the experiment. In Figure (a) we can see many loops with incorrect classification. However, when the wind direction is changed, all incorrect classifications disappear. 

\begin{table}[h!]
    \centering
    \caption{Maximum and mean Silhouette scores for all wind conditions scenarios in wind direction experiments.}
    \begin{tabular}{|c|c|c|}
    \hline
    Scenario & Maximum & Mean\\
    \hline 
        Light const. vs light shear - similar direction & 71.18 & 21.58 \\
        Light const. vs light shear - opposite direction & 92.16 & 90.33 \\
        \hline

    \end{tabular}
    \label{tab:wind direction_output}
\end{table}

\subsubsection{Cross-entropy method search strategy for thrust actions sequence}
\begin{figure}[!ht]
    \centering
    \includegraphics[width=0.28\textwidth]{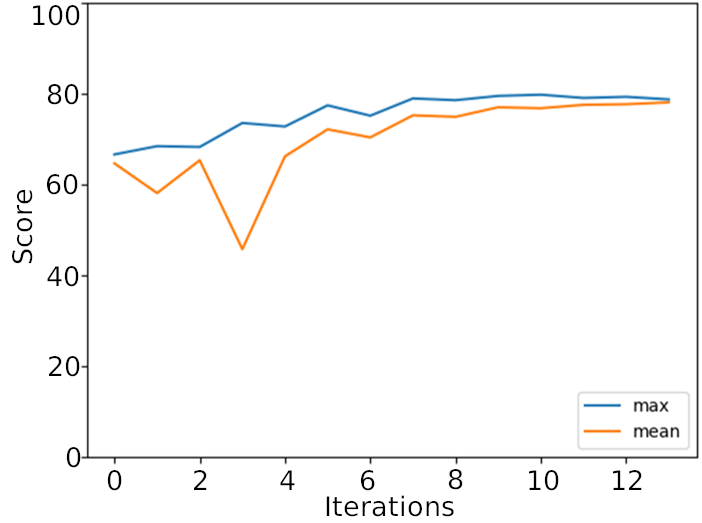}
    \caption{Silhouette score vs number of iterations for the light shear wind vs light constant wind case, using CEM with $\hat{\sigma}_0$=50 $N$=30 $N_e$=4.}
    \label{fig:CEM}
\end{figure}

In this case, the search method used for optimal UAV thrust sequences that maximise the identification of two wind conditions is the Cross-Entropy Method. Its algorithm is shown in section \ref{sec:method}. In Table \ref{tab:sim_sys_input_CEM} are shown the UAV, simulation and search method parameters. The thrust vector is defined by its magnitude and its three direction components. All of them are generated according to Gaussian distributions defined by $\mu$ and $\sigma$ which are updated with every iteration of the CEM method. In Figure \ref{fig:CEM} we can see the result for the light shear wind vs light constant wind case where $\hat{\sigma}_{mag,0} = 50$ and $\hat{\mu}_{mag,0}$ is centred at 25N for the thrust magnitude. For each thrust direction component, we started with $\hat{\sigma}_{dir,0} = 0.33$ and $\hat{\mu}_{dir,0} = 0$.  The number of samples is 30 for each iteration. The 4 elite samples ($N_e$) are the ones with the highest scores. These elite samples are used to calculate the new $\hat{\mu}_t$ and $\hat{\sigma}_t$ for the next iteration. 
An improvement trend can be seen through the iterations, as compared to the random choice of the sequences of the previous experiments. The graph plateaus around the 12th iteration. The max and mean results in Figure \ref{fig:CEM} can be compared directly with the ones in Figure \ref{fig:wind direction}(c) because they had the same wind condition. We can observe that the max value was 72.21 with the random search method while the max value with the CEM search reached 79.79. 

\begin{table}[h!]
    \centering
    \caption{Simulation, UAV and search strategy parameters for the CEM case.}
    \subfigure[UAV properties]{
    \begin{tabular}{|c|c|}
    \hline
        UAV mass (Kg) & 2 \\
        $C_D$ & 0.1 \\
        $S$ ($m^2$) & 0.01 \\
    \hline
    \end{tabular}}
    \subfigure[simulation  properties]{
    \begin{tabular}{|c|c|}
    \hline
        Time step (s) & 0.1 \\
        Total time  (s) & 12 \\
        Number of samples $N$   & 30 \\
        Elite samples $N_e$   & 4 \\
    \hline
    \end{tabular}}
    \subfigure[UAV thrust action properties ]{
    \begin{tabular}{|c|c|}
    \hline
        $F_{thrust}$ magnitude (N) & Gaussian distribution  \\
         & with $(\mu_{mag} , \sigma_{mag})$ \\
        $F_{thrust}$ directions [x y z] & Gaussian distribution  \\
         & with $(\vec{\mu}_{dir} , \vec{\sigma}_{dir})$\\
        $F_{thrust}$ application times (s) & Randomly selected \\
    \hline
    \end{tabular}}
    \label{tab:sim_sys_input_CEM}
\end{table}

\section{Discussion}
The results in Figure \ref{fig:wind speed} show that the lighter the wind the more difficult it is to identify it. The reason for this is that the forces acting on the UAV by the light wind are low and cause fewer changes in the trajectory which is the only input used in this method. We can also observe that some thrust actions give more useful information than others which results in better classification and better scores. This calls for a method for optimal thrust action sequences that maximise the classification. Here the Cross-Entropy Method is used.

The graphs in Figure \ref{fig:Range length} show that the larger the difference between wind speed ranges the easier it is to identify the two wind conditions, even if in this experiment it means reducing the number of environments.

Next, when testing the effect of the number of environments on the classification while keeping the range similarity constant, the increase in the number of environments increases the Silhouette score in all cases of Figure \ref{fig:Number of environments}. This is produced because the increase in the number of environments means more data input to the system. Nevertheless, the improvement plateaus around 20 environments, and for this reason the following tests were done using 20 environments.

It is important to note that the increase in the number of wind environments in each group could have improved the classification score in the previous experiment on the ranges difference, but it did not. Thus, the range length is a more sensitive parameter in classification than the number of wind environments in each group. 

As shown in Figure \ref{fig:thrust changes}, the number of UAV thrust vector changes during the simulation has no major effect on the identification. In contrast, Figure \ref{fig:wind direction} clearly shows that the difference in wind direction is a major factor for the identification. 

Figure \ref{fig:CEM} shows that the Cross-Entropy Method can improve the result for wind type identification. In the case of the experiment the improvement plateaus at the 12th iteration. 

Regarding future work, the collected observations in the current system are position versus time, but since observations are only required to be time series, future experimentation could be done with other UAV parameters such as attitude angles versus time or speed versus time and the actions could be roll, pitch, or yaw moments and not only the forces. Additionally, instead of using an open-loop approach where the thrust sequence is set at the beginning of the trajectory, a closed-loop system could be applied, observing the trajectory and producing thrust actions dynamically during the trajectory.

\section{Conclusions}
In this work we have applied a causal machine learning driven approach to identify the wind conditions of a UAV environment using only its position information. This method used for discerning causal factors in different environments, has been employed here in the aerospace domain for the first time. By applying this approach to the position trajectory of a UAV in diverse wind environments, we have forged a path to identifying wind conditions without the requisite of a dedicated wind speed sensor.

In this work we have utilised machine learning time series classification combined with the causal curiosity approach to discern various wind conditions, solely based on the UAV's trajectory. We have analysed three distinct wind environments: constant wind, shear wind, and turbulence. We analysed the effect of methodological and wind parameters in our approach and explored the use of the Cross-Entropy Method as a search strategy to maximise the accurate wind classification by employing optimal UAV manoeuvres.

The findings offer an enhanced understanding of UAV dynamics in wind disturbances and promise practical applications, such as optimising UAV pathways in windy scenarios for efficient energy consumption or faster destination reach.

 \bibliographystyle{ieeetr} 
 \bibliography{0_causal-learning}

\begin{thebibliography}{10}

\bibitem{abichandani2020wind}
P.~Abichandani, D.~Lobo, G.~Ford, D.~Bucci, and M.~Kam, ``Wind measurement and simulation techniques in multi-rotor small unmanned aerial vehicles,'' {\em IEEE Access}, vol.~8, pp.~54910--54927, 2020.

\bibitem{sontakke2020causal}
S.~A. Sontakke, A.~Mehrjou, L.~Itti, and B.~Sch{\"o}lkopf, ``Causal curiosity: Rl agents discovering self-supervised experiments for causal representation learning,'' in {\em International conference on machine learning}, pp.~9848--9858, PMLR, 2021.

\bibitem{MOURTZIS2021183}
D.~Mourtzis, J.~Angelopoulos, and N.~Panopoulos, ``Uavs for industrial applications: Identifying challenges and opportunities from the implementation point of view,'' {\em Procedia Manufacturing}, vol.~55, pp.~183--190, 2021.

\bibitem{ito2022load}
S.~Ito, K.~Akaiwa, Y.~Funabashi, H.~Nishikawa, X.~Kong, I.~Taniguchi, and H.~Tomiyama, ``Load and wind aware routing of delivery drones,'' {\em Drones}, vol.~6, no.~2, p.~50, 2022.

\bibitem{zeng2017energy}
Y.~Zeng and R.~Zhang, ``Energy-efficient uav communication with trajectory optimization,'' {\em IEEE Transactions on wireless communications}, vol.~16, no.~6, pp.~3747--3760, 2017.

\bibitem{cho2011wind}
A.~Cho, J.~Kim, S.~Lee, and C.~Kee, ``Wind estimation and airspeed calibration using a uav with a single-antenna gps receiver and pitot tube,'' {\em IEEE transactions on aerospace and electronic systems}, vol.~47, no.~1, pp.~109--117, 2011.

\bibitem{johansen2015estimation}
T.~A. Johansen, A.~Cristofaro, K.~S{\o}rensen, J.~M. Hansen, and T.~I. Fossen, ``On estimation of wind velocity, angle-of-attack and sideslip angle of small uavs using standard sensors,'' in {\em 2015 International Conference on Unmanned Aircraft Systems (ICUAS)}, pp.~510--519, IEEE, 2015.

\bibitem{borup2016nonlinear}
K.~T. Borup, T.~I. Fossen, and T.~A. Johansen, ``A nonlinear model-based wind velocity observer for unmanned aerial vehicles,'' {\em IFAC-PapersOnLine}, vol.~49, no.~18, pp.~276--283, 2016.

\bibitem{borup2019machine}
K.~T. Borup, T.~I. Fossen, and T.~A. Johansen, ``A machine learning approach for estimating air data parameters of small fixed-wing uavs using distributed pressure sensors,'' {\em IEEE Transactions on Aerospace and Electronic Systems}, vol.~56, no.~3, pp.~2157--2173, 2019.

\bibitem{thielicke2021towards}
W.~Thielicke, W.~H{\"u}bert, U.~M{\"u}ller, M.~Eggert, and P.~Wilhelm, ``Towards accurate and practical drone-based wind measurements with an ultrasonic anemometer,'' {\em Atmospheric Measurement Techniques}, vol.~14, no.~2, pp.~1303--1318, 2021.

\bibitem{perozzi2022using}
G.~Perozzi, D.~Efimov, J.-M. Biannic, and L.~Planckaert, ``Using a quadrotor as wind sensor: Time-varying parameter estimation algorithms,'' {\em International Journal of Control}, vol.~95, no.~1, pp.~126--137, 2022.

\bibitem{pearl2000models}
J.~Pearl, {\em Causality: Models, Reasoning, and Inference}.
\newblock Cambridge University Press, 2~ed., 2009.

\bibitem{glymour2016causal}
M.~Glymour, J.~Pearl, and N.~P. Jewell, {\em Causal inference in statistics: A primer}.
\newblock John Wiley \& Sons, 2016.

\bibitem{peters2017elements}
J.~Peters, D.~Janzing, and B.~Sch{\"o}lkopf, ``Elements of causal inference: Foundations and learning algorithms,'' 2017.

\bibitem{spirtes2000causation}
P.~Spirtes, C.~N. Glymour, R.~Scheines, and D.~Heckerman, {\em Causation, prediction, and search}.
\newblock MIT press, 2000.

\bibitem{scholkopf2022causality}
B.~Sch\"{o}lkopf, {\em Causality for Machine Learning}, p.~765–804.
\newblock New York, NY, USA: Association for Computing Machinery, 1~ed., 2022.

\bibitem{scholkopf2021toward}
B.~Sch{\"o}lkopf, F.~Locatello, S.~Bauer, N.~R. Ke, N.~Kalchbrenner, A.~Goyal, and Y.~Bengio, ``Toward causal representation learning,'' {\em Proceedings of the IEEE}, vol.~109, no.~5, pp.~612--634, 2021.

\bibitem{kaddour2022causal}
J.~Kaddour, A.~Lynch, Q.~Liu, M.~J. Kusner, and R.~Silva, ``Causal machine learning: A survey and open problems,'' {\em arXiv preprint arXiv:2206.15475}, 2022.

\bibitem{bareinboimcrlonline}
E.~Bareinboim, ``Towards causal reinforcement learning, icml tutorial,'' 2020.

\bibitem{zeng2023survey}
Y.~Zeng, R.~Cai, F.~Sun, L.~Huang, and Z.~Hao, ``A survey on causal reinforcement learning,'' {\em arXiv preprint arXiv:2302.05209}, 2023.

\bibitem{grimbly2021causal}
S.~J. Grimbly, J.~Shock, and A.~Pretorius, ``Causal multi-agent reinforcement learning: Review and open problems,'' {\em arXiv preprint arXiv:2111.06721}, 2021.

\bibitem{pathak2017curiosity}
D.~Pathak, P.~Agrawal, A.~A. Efros, and T.~Darrell, ``Curiosity-driven exploration by self-supervised prediction,'' in {\em International conference on machine learning}, pp.~2778--2787, PMLR, 2017.

\bibitem{burda2018large}
Y.~Burda, H.~Edwards, D.~Pathak, A.~Storkey, T.~Darrell, and A.~A. Efros, ``Large-scale study of curiosity-driven learning,'' {\em arXiv preprint arXiv:1808.04355}, 2018.

\bibitem{schmidhuber1991curious}
J.~Schmidhuber, ``Curious model-building control systems,'' in {\em Proceedings of the International Joint Conference on Neural Networks, Singapore, 1991}, vol.~2, pp.~1458--1463, IEEE press, 1991.

\bibitem{chentanez2004intrinsically}
N.~Chentanez, A.~Barto, and S.~Singh, ``Intrinsically motivated reinforcement learning,'' {\em Advances in neural information processing systems}, vol.~17, 2004.

\bibitem{oudeyer2009intrinsic}
P.-Y. Oudeyer and F.~Kaplan, ``What is intrinsic motivation? a typology of computational approaches,'' {\em Frontiers in neurorobotics}, p.~6, 2009.

\bibitem{still2012information}
S.~Still and D.~Precup, ``An information-theoretic approach to curiosity-driven reinforcement learning,'' {\em Theory in Biosciences}, vol.~131, pp.~139--148, 2012.

\bibitem{baldassarre2013intrinsically}
G.~Baldassarre and M.~Mirolli, {\em Intrinsically motivated learning in natural and artificial systems}.
\newblock Springer, 2013.

\bibitem{baranes2013active}
A.~Baranes and P.-Y. Oudeyer, ``Active learning of inverse models with intrinsically motivated goal exploration in robots,'' {\em Robotics and Autonomous Systems}, vol.~61, no.~1, pp.~49--73, 2013.

\bibitem{barto2013intrinsic}
A.~G. Barto, ``Intrinsic motivation and reinforcement learning,'' {\em Intrinsically motivated learning in natural and artificial systems}, pp.~17--47, 2013.

\bibitem{mohamed2015variational}
S.~Mohamed and D.~Jimenez~Rezende, ``Variational information maximisation for intrinsically motivated reinforcement learning,'' {\em Advances in neural information processing systems}, vol.~28, 2015.

\bibitem{forestier2017intrinsically}
S.~Forestier, R.~Portelas, Y.~Mollard, and P.-Y. Oudeyer, ``Intrinsically motivated goal exploration processes with automatic curriculum learning,'' {\em arXiv preprint arXiv:1708.02190}, 2017.

\bibitem{laversanne2018curiosity}
A.~Laversanne-Finot, A.~Pere, and P.-Y. Oudeyer, ``Curiosity driven exploration of learned disentangled goal spaces,'' in {\em Conference on Robot Learning}, pp.~487--504, PMLR, 2018.

\bibitem{oudeyer2018computational}
P.-Y. Oudeyer, ``Computational theories of curiosity-driven learning,'' {\em arXiv preprint arXiv:1802.10546}, 2018.

\bibitem{mcdonnell2024autonomous}
C.~McDonnell, M.~Arana-Catania, and S.~Upadhyay, ``Autonomous robotic arm manipulation for planetary missions using causal machine learning,'' in {\em 17th Symposium on Advanced Space Technologies in Robotics and Automation (ASTRA)}, 2023.

\bibitem{elkinton2006investigation}
M.~Elkinton, A.~Rogers, and J.~McGowan, ``An investigation of wind-shear models and experimental data trends for different terrains,'' {\em Wind Engineering}, vol.~30, no.~4, pp.~341--350, 2006.

\bibitem{moorhouse1982background}
D.~J. Moorhouse and R.~J. Woodcock, ``Background information and user guide for mil-f-8785c, military specification: Flying qualities of piloted airplanes,'' 1982.

\bibitem{muller2007dynamic}
M.~M{\"u}ller, ``Dynamic time warping,'' {\em Information retrieval for music and motion}, pp.~69--84, 2007.

\bibitem{JMLR:v21:20-091}
R.~Tavenard, J.~Faouzi, G.~Vandewiele, F.~Divo, G.~Androz, C.~Holtz, M.~Payne, R.~Yurchak, M.~Ru{\ss}wurm, K.~Kolar, {\em et~al.}, ``Tslearn, a machine learning toolkit for time series data,'' {\em The Journal of Machine Learning Research}, vol.~21, no.~1, pp.~4686--4691, 2020.

\bibitem{pedregosa2011scikit}
F.~Pedregosa, G.~Varoquaux, A.~Gramfort, V.~Michel, B.~Thirion, O.~Grisel, M.~Blondel, P.~Prettenhofer, R.~Weiss, V.~Dubourg, {\em et~al.}, ``Scikit-learn: Machine learning in python,'' {\em the Journal of machine Learning research}, vol.~12, pp.~2825--2830, 2011.

\bibitem{shahapure2020cluster}
K.~R. Shahapure and C.~Nicholas, ``Cluster quality analysis using silhouette score,'' in {\em 2020 IEEE 7th international conference on data science and advanced analytics (DSAA)}, pp.~747--748, IEEE, 2020.

\bibitem{botev2013cross}
Z.~I. Botev, D.~P. Kroese, R.~Y. Rubinstein, and P.~L’Ecuyer, ``The cross-entropy method for optimization,'' in {\em Handbook of statistics}, vol.~31, pp.~35--59, Elsevier, 2013.

\bibitem{kroese2006cross}
D.~P. Kroese, S.~Porotsky, and R.~Y. Rubinstein, ``The cross-entropy method for continuous multi-extremal optimization,'' {\em Methodology and Computing in Applied Probability}, vol.~8, pp.~383--407, 2006.

\bibitem{singhal2018unmanned}
G.~Singhal, B.~Bansod, and L.~Mathew, ``Unmanned aerial vehicle classification, applications and challenges: A review,'' {\em Preprints}, November 2018.

\bibitem{kurbanov2020development}
R.~Kurbanov and M.~Litvinov, ``Development of a gimbal for the parrot sequoia multispectral camera for the uav dji phantom 4 pro,'' {\em IOP Conference Series: Materials Science and Engineering}, vol.~1001, p.~012062, dec 2020.

\bibitem{polukhin2022development}
A.~A. Polukhin, M.~A. Litvinov, R.~K. Kurbanov, and S.~P. Klimova, ``Development of the parrot sequoia multispectral camera mount for the dji inspire 1 uav,'' in {\em Smart Innovation in Agriculture}, pp.~217--225, Springer, 2022.

\end{thebibliography}

\end{document}